# On the Size of the Online Kernel Sparsification Dictionary

Yi Sun                                                                          YI@IDSIA.CH
Faustino Gomez                                                                TINO@IDSIA.CH
Jürgen Schmidhuber                                                         JUERGEN@IDSIA.CH
IDSIA, USI & SUPSI, Switzerland

## Abstract

We analyze the size of the dictionary constructed from online kernel sparsification, using a novel formula that expresses the expected determinant of the kernel Gram matrix in terms of the eigenvalues of the covariance operator. Using this formula, we are able to connect the cardinality of the dictionary with the eigen-decay of the covariance operator. In particular, we show that under certain technical conditions, the size of the dictionary will always grow sublinearly in the number of data points, and, as a consequence, the kernel linear regressor constructed from the resulting dictionary is consistent.

## 1. Introduction

Kernel least squares (KLS) is a simple non-parametric regression method widely used in machine learning (e.g., see Schölkopf and Smola, 2002). Standard KLS requires storing and computing the (pseudo) inverse of the Gram matrix, and thus its complexity scales at least quadratically in the number of data points, rendering the method intractable for large data sets. In order to reduce the computational cost and avoid overfitting, it is common to replace the Gram matrix with a low rank approximation formed by projecting all *samples*[1] onto the span of a chosen subset or *dictionary* of samples. Using such an approximated Gram matrix can greatly reduce the computational cost of KLS, sometimes to linear in the number of data points. Generally speaking, there are two approaches to constructing the dictionary. The first is the *Nyström method* (Williams and Seeger, 2000), where a randomly selected subset is used. The second, which is the concern of this paper, is called *Online Kernel Sparsification* (OKS; Engel et al. 2004), where the dictionary is built up incrementally by incorporating new samples that cannot be represented well (in the least squares sense) using the current dictionary.

Since being proposed, OKS has found numerous applications in regression (Duy and Peters, 2010), classification (Slavakis et al., 2008) and reinforcement learning (Engel, 2005; Xu, 2006). Despite this empirical success, however, the theoretical understanding of OKS is still lacking. Most of the theoretical analysis has been done by Engel et al. (2004), who showed that the constructed dictionary is guaranteed to represent major fraction of the leading eigenvectors of the Gram matrix (Theorem 3.3, Engel et al. 2004). It was also proven that the dictionary stays finite if the set of possible samples is compact, and thus admits a finite covering number (Theorem 3.1, Engel et al. 2004). Yet, an important question remains open:

> How does the size of the dictionary scale with the number of samples if the set of possible samples does *not* admit a finite covering number, or if the covering number is too large compared to the size of the data set?

Answering this question allows us to: (1) estimate the computational complexity of OKS, and therefore the associated KLS method, more accurately, and (2) characterize the generalization capability of the KLS regression function obtained, as the usual risk bounds are controlled by the quotient between the size of the dictionary and the number of samples (e.g., see Györfi et al., 2004).

In this paper, we address this question theoretically. Our analysis proceeds in two steps:

1. We provide a novel formula expressing the *expected Gram determinant* over a set of i.i.d. samples in terms of the eigenvalues of the covariance operator. We then prove that the expected Gram

---

[1] With a little abuse of notation, we use *samples* to refer to elements in some reproducing kernel Hilbert space (RKHS), i.e., samples are the images of the feature map of the data points.

determinant diminishes with the cardinality of the set faster than any exponential function.

2. We observe that the Gram determinant over the OKS dictionary is lower bounded by some exponential function in the size of the dictionary. However, since step 1 concludes that the chance of a finding a big Gram matrix with large determinant is exceedingly small, the size of the dictionary must also stay small with high probability. Specifically, we show that the size of the dictionary will always grow sub-linearly in the number of data points, which implies consistency of KLS regressors constructed from the dictionary.

The rest of the paper is organized as follows: Section 2 describes the first step of our analysis, establishing a number of theoretical properties concerning the Gram determinant, including its expectation, decay, and moments. In section 3, we proceed to step 2, and analyze the growth of the size of the dictionary in OKS using the results from section 2. Section 4 briefly discusses the results and directions for future research.

## 2. The Determinant of a Gram matrix

Let $\mathcal{H}$ be a *separable* Hilbert space endowed with inner product $\langle \cdot, \cdot \rangle$, and $P$ be a distribution over $\mathcal{H}$. Assume $\mathbb{E}_{\phi \sim P} \|\phi\|^2 < \infty$, and let $\mathcal{C} = \mathbb{E}_{\phi \sim P} [\phi \otimes \phi]$ be the (non-centered) covariance operator, where $\otimes$ denotes the tensor product. Let $\lambda_1 \geq \lambda_2 \geq \cdots$ be the eigenvalues of $\mathcal{C}$ sorted in descending order, then $\sum \lambda_i < \infty$ (Theorem 2.1, Blanchard et al. 2007).

Given i.i.d. samples $\phi_1, \ldots, \phi_k \sim P$, let $G_k(\phi_{1:k})$ be[2] the $k \times k$ Gram matrix with $(i,j)$-th entry $\langle \phi_i, \phi_j \rangle$, and let $\det G_k$ be the determinant of $G_k$. Clearly, $\det G_k$ is a random variable from $\mathcal{H}^k$ to $\mathbb{R}$. Moreover, $\det G_k$ has bounded expectation since from Hadamard's inequality

$$0 \leq \mathbb{E}\left[\det G_k\right] \leq \mathbb{E}\left[\prod_{i=1}^k \|\phi_i\|^2\right] = \left(\mathbb{E}\|\phi\|^2\right)^k.$$

Let $\tilde{\lambda}_1 \geq \tilde{\lambda}_2 \geq \cdots \geq \tilde{\lambda}_k$ denote the eigenvalues of $k^{-1} G_k$ (and thus those of the empirical covariance operator $\tilde{\mathcal{C}}_k = k^{-1} \sum_{i=1}^k \phi_i \otimes \phi_i$) sorted in descending order. We assume the following condition.

**Assumption 1** $\lim_{k \to \infty} \sum_{i=1}^\infty \left|\tilde{\lambda}_i - \lambda_i\right| = 0$, *a.s.*, where we take $\tilde{\lambda}_i = 0$ for $i > k$.

The validity of this assumption will be discussed later in Section 4.1.

---

[2] We use $\phi_{1:k}$ as a short hand for $\phi_1, \ldots, \phi_k$.

### 2.1. A Formula for the Expectation of the Gram Determinant

Before presenting our first main result (Theorem 1), we introduce some additional notation. The *elementary symmetric polynomial*[3] of order $k$ over $n$ variables is defined as

$$\nu_{n,k}(\lambda_{1:n}) = k! \sum_{1 \leq i_1 < i_2 < \cdots < i_k \leq n} \lambda_{i_1} \lambda_{i_2} \cdots \lambda_{i_k},$$

where the summation runs over all $k$-subsets of $\{1, \ldots, n\}$. We denote the infinite extension of $\nu_{n,k}$ as

$$\nu_k(\lambda_1, \lambda_2, \ldots) = k! \sum_{1 \leq i_1 < i_2 < \cdots < i_k} \lambda_{i_1} \lambda_{i_2} \cdots \lambda_{i_k},$$

whenever the infinite sum exists. For simplicity, $\nu_k$ and $\nu_{n,k}$ denote both the function and their respective values with default argument $(\lambda_1, \lambda_2, \ldots)$, and we only write down the arguments when they differ from $(\lambda_1, \lambda_2, \ldots)$. Some of the useful properties of $\nu_{n,k}$ and $\nu_k$ are summarized in the following Lemma.

**Lemma 1** *We have*

a) $\nu_{n,k} \geq \nu_{n-1,k} \geq 0$, *and* $\lim_{n \to \infty} \nu_{n,k} = \nu_k$.

b) $\nu_{n,k} = k \lambda_n \nu_{n-1,k-1} + \nu_{n-1,k}$,

c) $\nu_k^2 \geq \nu_{k-1} \nu_{k+1}$ *(Newton's inequality)*,

d) $\nu_k^{\frac{1}{k}} \geq \nu_{k+1}^{\frac{1}{k+1}}$ *(Maclaurin's inequality)*.

**Proof.** We only prove the limit in a) exists. The other properties can be derived easily using the limit argument and the properties of elementary symmetric polynomials (e.g., see Niculescu, 2000). In particular, c) is a direct consequence of Newton's inequality, and d) is a rephrase of Maclaurin's inequality.

Note that $\nu_{n,k}$ is a non-decreasing sequence of $n$. Moreover,

$$\nu_{n,k} = k! \sum_{1 \leq i_1 < i_2 < \cdots < i_k \leq n} \lambda_{i_1} \cdots \lambda_{i_k} < k! \left(\sum_{i=1}^n \lambda_i\right)^k$$

is bounded because $\sum \lambda_i < \infty$. Therefore the limit exists. ∎

Note that property b) enables us to compute $\nu_{n,k}$ in $O(nk)$ time using dynamic programming. More precisely, this is done by initializing i) $\nu_{1,1} = \lambda_1$, ii) $\nu_{i,1} = \nu_{i-1,1} + \lambda_i$ for $i = 1, \ldots, n$, and iii) $\nu_{i,i} = i \lambda_i \nu_{i-1,i-1}$ for $i = 1, \ldots, k$, and then applying the recursion in b).

The following theorem gives an explicit representation of the expectation of $\det G_k$ in terms of the eigenvalues of $\mathcal{C}$.

---

[3] Note that the standard definition does not have the $k!$ term.

**Theorem 1** $\mathbb{E}\left[\det G_k\left(\phi_{1:k}\right)\right] = \nu_k$

That is, the expectation of the determinant of a Gram matrix built from $k$ samples is equal to the $k-th$ order elementary symmetric polynomial over the eigenvalues of the covariance operator.

**Proof.** [4] Let $\phi_1, \ldots, \phi_n \sim P$, and $G_n = G_n(\phi_{1:n})$ be the corresponding Gram matrix. Denote $\tilde{\lambda}_1, \ldots, \tilde{\lambda}_n$ the eigenvalues of $n^{-1}G_n$, so that $n\tilde{\lambda}_i$ are the eigenvalues of $G_n$. The characteristic polynomial of $G_n$ is given by $f(\lambda) = \det(G_n - \lambda I)$. By definition,

$$f(-\lambda) = \prod_{i=1}^n (\lambda + n\lambda_i)$$

$$= \sum_{k=0}^n n^k \left(\sum_{1 \leq i_1 < \cdots < i_k \leq n} \lambda_{i_1} \cdots \lambda_{i_k}\right) \cdot \lambda^{n-k}$$

$$= \sum_{k=0}^n n^k \frac{\nu_{n,k}\left(\tilde{\lambda}_{1:n}\right)}{k!} \cdot \lambda^{n-k}.$$

Alternatively, we can express $f(-\lambda)$ using the determinants of the principal submatrices (see for example Meyer, 2001, pp.494), which are Gram matrices by themselves:

$$f(-\lambda) = \sum_{k=0}^n \sum_{\mathcal{I} \in [n]_k} \det G_k(\phi_\mathcal{I}) \cdot \lambda^{n-k},$$

where $[n]_k$ is the family of $k$-subsets in $\{1, \ldots, n\}$, and $\phi_\mathcal{I}$ denotes $\{\phi_i\}_{i \in \mathcal{I}}$. Divide the coefficients before $\lambda^{n-k}$ by binomial coefficient $\binom{n}{k}$ to get the identity:

$$\binom{n}{k}^{-1} \sum_{\mathcal{I} \in [n]_k} \det G_k(\phi_\mathcal{I}) = \frac{(n-k)!n^k}{n!}\nu_{n,k}\left(\tilde{\lambda}_{1:n}\right).$$

The l.h.s. is a U-statistic (Serfling, 1980) with kernel $\det G_k$. Since $\mathbb{E}[\det G_k] < \infty$, the law of large numbers for U-statistics (Hoeffding, 1961) asserts that

$$\mathbb{E}[\det G_k] = \lim_{n \to \infty} \binom{n}{k}^{-1} \sum_{\mathcal{I} \in [n]_k} \det G_k(\phi_\mathcal{I}), \text{ a.s.}$$

Now consider the r.h.s. For the first term

$$\lim_{n \to \infty} \frac{(n-k)!n^k}{n!} = \lim_{n \to \infty} \frac{n}{n-1} \cdots \frac{n}{n-k+1} = 1.$$

---
[4] An alternative proof may be derived using the generator function of $\mathbb{E}[\det G_k]$ (Martin, 2007). Unfortunately, the result is only briefly alluded to in the slides, and no detailed documentation has been made available up to now.

For the second term, we have

$$\nu_{n,k}\left(\tilde{\lambda}_{1:n}\right) - \nu_k$$
$$= \sum_{i=1}^n \left(\nu_{n,k}\left(\lambda_{1:i-1}, \tilde{\lambda}_{i:n}\right) - \nu_{n,k}\left(\lambda_{1:i}, \tilde{\lambda}_{i+1:n}\right)\right)$$
$$+ \left(\nu_{n,k}(\lambda_1, \ldots, \lambda_n) - \nu_k\right).$$

Note that

$$\left|\nu_{n,k}\left(\lambda_{1:i-1}, \tilde{\lambda}_{i:n}\right) - \nu_{n,k}\left(\lambda_{1:i}, \tilde{\lambda}_{i+1:n}\right)\right|$$
$$= \left|k\left(\tilde{\lambda}_i - \lambda_i\right)\nu_{n-1,k-1}\left(\lambda_{1:i-1}, \tilde{\lambda}_{i+1:n}\right)\right|$$
$$\leq k\left|\tilde{\lambda}_i - \lambda_i\right|\nu_{n,k-1}\left(\lambda_{1:i-1}, \lambda_i, \tilde{\lambda}_{i+1:n}\right)$$
$$\leq k\left|\tilde{\lambda}_i - \lambda_i\right|\nu^*_{n,k-1},$$

where

$$\nu^*_{n,k-1} = \nu_{n,k-1}\left(\max\left\{\tilde{\lambda}_1, \lambda_1\right\}, \ldots, \max\left\{\tilde{\lambda}_n, \lambda_n\right\}\right)$$

is bounded as $\sum \max\left\{\tilde{\lambda}_i, \lambda_i\right\} < \infty$. Therefore

$$\left|\nu_{n,k}\left(\tilde{\lambda}_{1:n}\right) - \nu_k\right|$$
$$\leq k\nu^*_{n,k-1}\sum_{i=1}^n \left|\tilde{\lambda}_i - \lambda_i\right| + \left|\nu_{n,k}(\lambda_1, \ldots, \lambda_n) - \nu_k\right|$$
$$\to 0, \text{ a.s.}$$

The first summand vanishes because of Assumption 1, and the second one diminishes because of Lemma 1 a). As a result,

$$\lim_{n \to \infty} \frac{(n-k)!n^k}{n!}\nu_{n,k}\left(\tilde{\lambda}_{1:n}\right) = \nu_k.$$

∎

### 2.2. The Decaying Speed of $\mathbb{E}[\det G_k]$

It is not immediately obvious how $\nu_k = \mathbb{E}[\det G_k]$ behaves with increasing $k$. Here we provide a direct link between the speed with which $\nu_k$ approaches zero and the tail behavior of $\{\lambda_i\}$. The analysis is based on the following lemma.

**Lemma 2** Let $\lambda^{(0)} = \sum \lambda_j$, and $\lambda^{(k)} = \sum_{j>k} \lambda_j$. Then

$$\log \nu_{k+s} - \log \nu_k \leq s \log \lambda^{(k)} + \log \binom{k+s}{k}.$$

**Proof.** Note that

$$\nu_{k+s} = \frac{(k+s)!}{k!s!}k!\sum_{1 \leq i_1 < \cdots < i_k} \lambda_{i_1} \cdots \lambda_{i_k} \cdot s!\sum_{i_k < j_1 < \cdots j_s} \lambda_{j_1} \cdots \lambda_{j_s}$$

$$= \binom{k+s}{k}k!\sum_{1 \leq i_1 < \cdots < i_k} \lambda_{i_1} \cdots \lambda_{i_k} \cdot \nu_s(\lambda_{i_k+1}, \lambda_{i_k+2}, \ldots)$$

Since $\lambda_i$ is decreasing and $i_k \geq k$, we have for all $i_k$

$$\nu_s(\lambda_{i_k+1}, \lambda_{i_k+2}, \dots) \leq \nu_s(\lambda_{k+1}, \lambda_{k+2}, \dots)$$
$$\leq \left(\sum_{j>k} \lambda_j\right)^s,$$

where the last inequality is from Lemma 1 d). Therefore,

$$\nu_{k+s} \leq \binom{k+s}{k} \left(\lambda^{(k)}\right)^s \cdot \nu_k.$$

Taking the logarithm gives the desired result. ∎

An immediate consequence is that $\nu_k$ converges to 0 faster than any exponential function.

**Corollary 1** *For any $\alpha > 0$, $\lim_{k\to\infty} \alpha^{-k} \nu_k = 0$.*

**Proof.** Assume $k$ is fixed and $s$ is large. From Stirling's formula

$$\log \binom{k+s}{k} = k \log\left(1 + \frac{s}{k}\right) + s \log\left(1 + \frac{k}{s}\right)$$
$$+ O(\log s) < s + O(\log s),$$

where we use $\log(1+x) < x$ for all $x > -1$.
By Lemma 2,

$$\log \nu_{k+s} - (k+s) \log \alpha$$
$$\leq s\left[1 - \log \alpha + \log \lambda^{(k)}\right] + O(\log s).$$

Since $\sum \lambda_i < \infty$, we can pick a $k^*$ such that $\log\left(\sum_{j>k^*} \lambda_j\right) < -2 + \log \alpha$, then

$$\lim_{k\to\infty} \log \nu_k - k \log \alpha = \lim_{s\to\infty} (\log \nu_{k^*+s} - (k^* + s) \log \alpha)$$
$$< \lim_{s\to\infty} (-s + O(\log s)) = -\infty,$$

and thus $\lim_{k\to\infty} \alpha^{-k} \nu_k = 0$. ∎

**Remark 1** *We can also bound $\nu_k$ in terms of $\lambda^{(k)}$ using Lemma 2. For exponential decay, i.e., $\lambda_i \sim O(\sigma^{-i})$, we take $s = 1$, then*

$$\log \nu_k < -\frac{k^2}{2} \log \sigma + \log k! + O(k).$$

*The bound is tight since for $\lambda_i = \sigma^{-i}$, direct computation gives*

$$\nu_k = k! \sum_{i_1=1}^{\infty} \lambda_{i_1} \cdots \sum_{i_k=i_{k-1}+1}^{\infty} \lambda_{i_k} = \frac{k! \sigma^{-k}}{\prod_{i=1}^{k}(\sigma^i - 1)}.$$

*Taking the logarithm and applying some algebra we get*

$$\log \nu_k = -\frac{k^2}{2} \log \sigma + \log k! + O(k).$$

*For polynomial decay, i.e., $\lambda_i \sim O(i^{-(1+p)})$, $\sum_{i \geq k} i^{-(1+p)} \sim \frac{k^{-p}}{p}$, we set $s = k$, then*

$$\log \nu_{2k} - \log \nu_k \leq k \log \lambda^{(k)} + \log \binom{2k}{k}.$$

*Using Stirling's formula,*

$$\log(2k)! - 2 \log k! = k \log 4 + O(\log k).$$

*Therefore,*

$$\log \frac{\nu_{2k}}{\nu_k} \leq -pk \log k + k \log \frac{4}{p} + O(\log k),$$

*which characterizes the convergence of $\nu_k$.*

### 2.3. Bounding the Moments of the Gram Determinant

In this section we prove a simple result concerning the moment $\mathbb{E}[(\det G_k)^m]$, with the additional assumption that $\mathcal{H}$ is the reproducing kernel Hilbert space (RKHS) associated with some bounded Mercer kernel $\ell(x, x')$. Note that for any $m \geq 1$, $\ell^{(m)}(x, x') = (\ell(x, x'))^m$ is still a bounded Mercer kernel. Let $\mathcal{H}^{(m)}$ be the RKHS associated with $\ell^{(m)}$ and denote $\lambda_1^{(m)} \geq \lambda_2^{(m)} \geq \cdots$ the eigenvalues of the corresponding covariance operator in $\mathcal{H}^{(m)}$. We have the following bound.

**Theorem 2** $\mathbb{E}[(\det G_k)^m] \leq \nu_k\left(\lambda_1^{(m)}, \lambda_2^{(m)}, \dots\right)$ *for $m = 2, 3, \dots$.*

**Proof.** Let $A \circ B$ be the Hadamard product of $A$ and $B$. We use the well-known fact: If $A, B$ are positive semi-definite, then

$$\det(A \circ B) \geq \det(A) \det(B).$$

Repeating the process in the proof of Theorem 1, and applying

$$\det G_k^{(m)} \geq (\det G_k)^m$$

gives the result. ∎

**Remark 2** *Theorem 2 allows us to estimate empirically the bound of $\mathbb{E}[(\det G_k)^m]$ without enumerating all subsets of size $k$. Moreover, for RBF and polynomial kernels, $\ell^{(m)}$ stays RBF and polynomial, respectively. However, it remains unknown how $\lambda_i^{(m)}$ behaves in the general case.*

## 3. Analyzing Online Kernel Sparsification

In OKS, the dictionary $\mathcal{D}$ is initially empty. When a new sample $\phi$ arrives[5], it is added to the dictionary if

$$\frac{\det G_{\mathcal{D} \cup \{\phi\}}}{\det G_{\mathcal{D}}} > \alpha,$$

where $G_{\mathcal{D}}$ and $G_{\mathcal{D} \cup \{\phi\}}$ are the respective Gram matrices of $\mathcal{D}$ and $\mathcal{D} \cup \{\phi\}$, and $\alpha > 0$ is a user-defined constant controlling the approximation error. Note that our notation is equivalent to the form originally proposed by Engel et al. (2004) as

$$\frac{\det G_{\mathcal{D} \cup \{\phi\}}}{\det G_{\mathcal{D}}} = \langle \phi, \phi \rangle - g^\top G_{\mathcal{D}}^- g = \min_{\psi \in \text{span } \mathcal{D}} \|\phi - \psi\|^2$$

where $g = [\langle \phi, \phi_1 \rangle, \ldots, \langle \phi, \phi_{|\mathcal{D}|} \rangle]^\top$ for $\mathcal{D} = \{\phi_1, \ldots, \phi_{|\mathcal{D}|}\}$, and $G_{\mathcal{D}}^-$ is the inverse of $G_{\mathcal{D}}$. The new $\phi$ can be added in $O(|\mathcal{D}|^2)$ time if $G_{\mathcal{D}}^-$ is updated incrementally, for a total computational complexity $O(|\mathcal{D}|^2 n)$ for $n$ samples.

Our analysis is based on the key observation that

$$\det G_{\mathcal{D}} > \alpha^{|\mathcal{D}|}.$$

Since we have shown in the previous section that $\alpha^{-k} \mathbb{E}[\det G_k] \to 0$, the chance of finding a subset with the property that $\det G_{\mathcal{D}} > \alpha^{|\mathcal{D}|}$ will diminish as $|\mathcal{D}|$ grows, making a large dictionary unlikely.

More specifically, let $\phi_1, \ldots, \phi_n$ be $n$ i.i.d. samples from $P$, and let $\mathcal{D}_n$ be the dictionary constructed from $\phi_{1:n}$. Denote $[n]_k$ to be the family of all $k$-subsets of $\{1, \ldots, n\}$. For $\mathcal{A} \in [n]_k$, let

$$\rho_k(\phi_{\mathcal{A}}) = \mathbb{I}\left[\det G_k(\phi_{\mathcal{A}}) > \alpha^k\right],$$

where $\mathbb{I}[\cdot]$ is the indicator function. Define

$$k_n^* = \operatorname*{argmax}_k \left\{ \sum_{\mathcal{A} \in [n]_k} \rho_k(\phi_{\mathcal{A}}) > 0 \right\}.$$

Then clearly $|\mathcal{D}_n| < k_n^*$, and we may study $k_n^*$ instead of $|\mathcal{D}|$. Intuitively, $k_n^*$ characterizes the dimensionality of the linear space spanned by $\phi_{1:n}$, because for any subset larger than $k_n^*$ there will be some $\phi$ which can be represented within error $\alpha$ by the linear combination of $\phi_{1:n}$.

To characterize $k_n^*$ we study $\mathbb{P}[k_n^* \geq k]$. The following lemma shows that this probability is equal to the probability of the existence of $k$-subsets $\mathcal{A}$ with $\rho_k(\mathcal{A}) = 1$.

---

[5] In practice $\phi$ are often features in some RKHS induced by a kernel, and we store samples in the original domain. However we assume $\mathcal{D}$ is made of features for conceptual simplicity.

**Lemma 3** $\mathbb{P}[k_n^* \geq k] = \mathbb{P}\left[\sum_{\mathcal{A} \in [n]_k} \rho_k(\phi_{\mathcal{A}}) > 0\right].$

**Proof.** By definition

$$\mathbb{P}\left[\sum_{\mathcal{A} \in [n]_k} \rho_k(\phi_{\mathcal{A}}) > 0\right] \leq \mathbb{P}\left[\bigcup_{k' \geq k} \left\{ \sum_{\mathcal{A} \in [n]_{k'}} \rho_{k'}(\phi_{\mathcal{A}}) > 0 \right\}\right]$$
$$= \mathbb{P}[k_n^* \geq k].$$

Therefore the equality is not trivial.

From Theorem 5 in Cover and Thomas (1988),

$$\left(\frac{\det G_{k+1}(\phi_{1:k+1})}{\alpha^{k+1}}\right)^{\frac{1}{k+1}} \leq \frac{1}{k+1} \sum_{\mathcal{A} \in [k+1]_k} \left(\frac{\det G_k(\phi_{\mathcal{A}})}{\alpha^k}\right)^{\frac{1}{k}}.$$

Therefore, $\sum_{\mathcal{A} \in [n]_k} \rho_k(\phi_{\mathcal{A}}) = 0$ implies $\sum_{\mathcal{A} \in [n]_{k'}} \rho_{k'}(\phi_{\mathcal{A}}) = 0$ for all $k' \geq k$, and thus

$$\mathbb{P}\left[\sum_{\mathcal{A} \in [n]_k} \rho_k(\phi_{\mathcal{A}}) = 0\right] \leq \mathbb{P}\left[\bigcap_{k' \geq k} \left\{ \sum_{\mathcal{A} \in [n]_{k'}} \rho_{k'}(\phi_{\mathcal{A}}) = 0 \right\}\right].$$

Taking the complement on both sides,

$$\mathbb{P}\left[\sum_{\mathcal{A} \in [n]_k} \rho_k(\phi_{\mathcal{A}}) > 0\right] \geq \mathbb{P}\left[\bigcup_{k' \geq k} \left\{ \sum_{\mathcal{A} \in [n]_{k'}} \rho_{k'}(\phi_{\mathcal{A}}) > 0 \right\}\right]$$
$$= \mathbb{P}[k_n^* \geq k].$$

∎

We may now proceed to bound $k_n^*$, using basic tools from probability theory.

**Theorem 3** $\mathbb{P}[|\mathcal{D}_n| \geq k] \leq \mathbb{P}[k_n^* \geq k] < \alpha^{-k} \binom{n}{k} \nu_k.$

**Proof.** Note that

$$\mathbb{E}\left[\sum_{\mathcal{A} \subset [n]_k} \rho_k(\phi_{\mathcal{A}})\right] = \binom{n}{k} \mathbb{E}[\rho_k] = \binom{n}{k} \mathbb{P}\left[\det G_k > \alpha^k\right].$$

From Markov's inequality,

$$\mathbb{P}\left[\det G_k > \alpha^k\right] < \frac{\mathbb{E}[\det G_k]}{\alpha^k}.$$

It then follows

$$\mathbb{P}[k_n^* \geq k] = \mathbb{P}\left[\sum_{\mathcal{A} \in [n]_k} \rho_k(\phi_{\mathcal{A}}) \geq 1\right]$$
$$\leq \mathbb{E}\left[\sum_{\mathcal{A} \subset [n]_k} \rho_k(\phi_{\mathcal{A}})\right] < \binom{n}{k} \frac{\mathbb{E}[\det G_k]}{\alpha^k}.$$

Here we use the fact that $\rho_k$ is $\{0, 1\}$-valued, and apply Markov's inequality again. ∎

Note that the proof only uses Markov's inequality, which usually provides bounds that are by no means tight. The possibility of strengthening the bound is discussed in the next section. However, even with this simple analysis, some interesting results for the size of $\mathcal{D}$ can be obtained. The first is the following corollary.

**Corollary 2** *For any $\varepsilon \in (0, 1]$,*

$$\lim_{n \to \infty} \mathbb{P}\left[\frac{k_n^*}{n} \geq \varepsilon\right] = 0.$$

**Proof.** For simplicity assume $\varepsilon n$ is an integer. Let $k = n\varepsilon$, then

$$\binom{n}{k}\frac{\nu_k}{\alpha^k} = \binom{\varepsilon^{-1}k}{k}\frac{\nu_k}{\alpha^k}.$$

Using Stirling's formula,

$$\log \binom{\varepsilon^{-1}k}{k} = \log(\varepsilon^{-1}k)! - \log k! - \log\left((\varepsilon^{-1}-1)k\right)!$$
$$= \gamma k + O(\log k),$$

where

$$\gamma = \frac{1}{\varepsilon}\log\frac{1}{\varepsilon} - \left(\frac{1}{\varepsilon} - 1\right)\log\left(\frac{1}{\varepsilon} - 1\right).$$

Therefore, following Corollary 1 and Theorem 3,

$$\lim_{n\to\infty} \mathbb{P}\left[\frac{k_n^*}{n} \geq \varepsilon\right] < \lim_{n\to\infty, k=\varepsilon n} \binom{n}{k}\frac{\mathbb{E}[\det G_k]}{\alpha^k}$$
$$= \lim_{k\to\infty} \binom{\varepsilon^{-1}k}{k}\frac{\nu_k}{\alpha^k} = 0.$$

∎

**Remark 3** *By definition, Corollary 2 indicates that $n^{-1}k_n^* \to 0$ in probability, or the size of the dictionary grows only sub-linearly with the number of samples. Assuming finite variance of the response variable, it immediately follows that the ordinary linear regressor constructed using features obtained from OKS is consistent, as the generalization error is controlled by $n^{-1}|\mathcal{D}|$ (e.g., see Györfi et al., 2004).*

The next corollary provides a bound given a finite number of samples.

**Corollary 3** *For arbitrary $\delta > 0$ and*

$$n < \frac{\alpha k}{e}\left(\frac{\delta}{\nu_k}\right)^{\frac{1}{k}},$$

*we have $\mathbb{P}[|\mathcal{D}_n| > k] < \delta$.*

**Remark 4** *It is possible to give a bound in $k$ rather than $n$. However, such a bound requires the inversion of $\nu_k$ and complicates the notation.*

**Proof.** Assume $n = \varepsilon^{-1}k$. Rewrite Theorem 3 as

$$\mathbb{P}\left[|\mathcal{D}_{\varepsilon^{-1}k}| \geq k\right] < \alpha^{-k}\binom{\varepsilon^{-1}k}{k}\nu_k.$$

Using the simple relation $\binom{\varepsilon^{-1}k}{k} < \left(\varepsilon^{-1}e\right)^k$, we have

$$\log \alpha^{-k}\binom{\varepsilon^{-1}k}{k}\nu_k < k(1 - \log \alpha) - k\log\varepsilon + \log \nu_k.$$

Letting the r.h.s. equal $\log \delta$, it follows that

$$\log\frac{e}{\alpha} + \frac{1}{k}\log\frac{\nu_k}{\delta} = \log \varepsilon, \text{ and } \varepsilon = \frac{e}{\alpha}\left(\frac{\nu_k}{\delta}\right)^{\frac{1}{k}}.$$

∎

Using Corollary 3, an upper bound on the dictionary size can be derived using $\{\lambda_i\}$, and the impact of $\alpha$ on the dictionary size can be analyzed.

From the previous discussion, if $\lambda_i \leq \sigma^{-i}$, then

$$\nu_k \leq \frac{k!\,(\sigma)^{-k}}{\prod_{i=1}^k(\sigma^i - 1)} = \frac{k!\,(\sigma)^{-k}}{\sigma^{\frac{k(k+1)}{2}}\prod_{i=1}^k(1 - \sigma^{-i})},$$

and some elementary manipulation gives

$$\sum_{i=1}^k \log\left(1 - \sigma^{-i}\right) > -\frac{\sigma}{(\sigma-1)^2}.$$

Therefore,

$$\frac{1}{k}\log \nu_k > -\frac{k}{2}\log\sigma + \log k - \frac{3}{2} - \log\sigma - \frac{1}{k}\frac{\sigma}{(\sigma-1)^2}.$$

Plugging this into Corollary 3, $n < \frac{\alpha}{\beta}\delta^{\frac{1}{k}}\sigma^{\frac{k}{2}}$, where $\beta$ is some constant depending on $\sigma$, which implies $k \sim O(\log(n))$. Similarly, for polynomial decay $n^{-(1+p)}$, we have for large $k$

$$\frac{1}{k}\log\frac{\nu_{2k}}{\nu_k} < -p\log k + \log\frac{4}{p},$$

and then $n > \alpha\delta^{\frac{1}{k}}k^{1+p}$. Therefore, the dictionary size grows approximately at the rate of $n^{\frac{1}{1+p}}$. Note that the order of magnitude of these bounds coincides with the number of eigenvalues above certain threshold (Bach and Jordan, 2002, Table 3).

## 4. Discussion

This paper presented a rigorous theoretical analysis of how the dictionary in online kernel sparsification scales with respect to the number of samples, based on properties of the Gram matrix determinant. This work should lead to a better understanding of OKS, both in terms of its computational complexity, and the generalization capabilities associated with kernel least squares regressors. Three additional points are discussed below concerning a) the validity of Assumption 1, b) how our results relate to the Nyström method, and c) how the analysis can be potentially developed further.

### 4.1. On Assumption 1

Under the mild condition $\mathbb{E}_{\phi \sim P} \|\phi\|^2 < \infty$, it can be seen that $\langle \cdot, \cdot \rangle$ is a Mercer kernel in the sense of Definition 2.15 in Braun (2005), and subsequently by Theorem 3.26 therein, it follows that the $\delta_2$ distance (Koltchinskii and Giné 2000, pp. 116) between $\left\{\tilde{\lambda}_i\right\}$ and $\{\lambda_i\}$ vanishes almost surely.

However, the convergence of the spectrum in $\delta_2$ metric is insufficient for Theorem 1 to hold, and the stronger $L1$ convergence of the eigenspectrum is needed. It is possible to drop Assumption 1 altogether and base the discussion on $\lim_{n \to \infty} \nu_{n,k}$ instead, where the limit always exists and equals to $\mathbb{E}[\det G_k]$. Otherwise, following the analysis by Gretton et al. (2009), we may provide sufficient conditions[6] to Assumption 1 using the following extension of the Hoffman–Wielandt inequality (Theorem 3, Bhatia and Elsner 1994)

$$\sum_i \left|\tilde{\lambda}_i - \lambda_i\right| \leq \left\|\tilde{\mathcal{C}}_k - \mathcal{C}\right\|_1,$$

where $\|\cdot\|_1$ denotes the trace norm. Using Proposition 12 in Harchaoui et al. (2008), the convergence of $\left\|\tilde{\mathcal{C}}_k - \mathcal{C}\right\|_1$ to zero can be established provided that i) $\mathcal{H}$ is a *separable* RKHS (e.g., an RKHS induced by a continuous kernel over a separable metric space; Steinwart et al. 2006) induced by some *bounded* kernel, and ii) the eigenspectrum of $\mathcal{C}$ satisfies $\sum_i \lambda_i^{\frac{1}{2}} < \infty$.

### 4.2. Comparison with Nyström Method

A similar approach to OKS for reducing the computational cost of kernel methods is the Nyström method (Williams and Seeger, 2000), where the dictionary consists of a subset of samples chosen at random. One distinction of the two methods, following from the analysis before, is that the dictionary from OKS satisfies $\det G_\mathcal{D} > \alpha^{|\mathcal{D}|}$, while the randomly selected subset $\tilde{\mathcal{D}}$, satisfies $\det G_{\tilde{\mathcal{D}}} \ll \alpha^{|\tilde{\mathcal{D}}|}$ for larger $\mathcal{D}$. Therefore,

$$\frac{\det G_n}{\det G_\mathcal{D}} \ll \frac{\det G_n}{\det G_{\tilde{\mathcal{D}}}}.$$

---

[6]We thank the anonymous reviewers for pointing this out.

From an information theoretic point of view, $\log \frac{\det G_n}{\det G_{\tilde{\mathcal{D}}}}$ can be interpreted as the conditional entropy (Cover and Thomas, 1988), which indicates that $\tilde{\mathcal{D}}$ captures less information about the data sets.

The theoretical study of the Nyström method by Drineas and Mahoney (2005) suggests that $O\left(\alpha^{-4}k\right)$ samples are needed to approximate the first $k$ eigenvectors well, which is linear in $k$, irrespective of the sample size. A recent study (Jin et al., 2012) shows that assuming bounded kernel, the spectral norm of the approximation error between the true and the approximated Gram matrix scales at a rate of $O\left(n|\mathcal{D}|^{-\frac{1}{2}}\right)$, and in the case of $\lambda_i \sim i^{-p}$, an $O\left(n|\mathcal{D}|^{1-p}\right)$ rate may be obtained. In contrast, the results in this paper are over the dictionary size $|\mathcal{D}|$, and the approximation error is controlled by $\alpha$. In particular, assuming bounded kernel, the $(i,j)$-th entry of the difference between the true and approximated Gram matrix using OKS is bounded by

$$|\langle \phi_i, \phi_j \rangle - \langle \Pi_\mathcal{D} \phi_i, \Pi_\mathcal{D} \phi_j \rangle| < 2 \sup \|\phi\| \sqrt{\alpha},$$

where $\Pi_\mathcal{D}$ denotes the projection operator into the space spanned by $\mathcal{D}$ and the inequality follows from the Cauchy-Schwartz inequality. Using the fact that $\|A\|_2 \leq \sqrt{\|A\|_1 \|A\|_\infty}$ for arbitrary matrix $A$, where $\|\cdot\|_2$, $\|\cdot\|_1$ and $\|\cdot\|_\infty$ respectively denote the spectral norm, maximum absolute column sum norm and maximum absolute row sum norm, we conclude that the spectral norm of the approximation error is controlled by $O(n\sqrt{\alpha})$, which is a non-probabilistic bound and does not explicitly depend on the dictionary size.

### 4.3. On Strengthening the Bound

The proof of Theorem 2 uses Markov's inequality to bound both $\mathbb{P}[\det G_k > \alpha^k]$, and the probability of $\sum_{\mathcal{A} \in [n]_k} \rho_k(\mathcal{A}) \neq 0$. In practice, this bound is hardly satisfying. One possibility is to strengthen the bound by incorporating information from higher order moments (Philips and Nelson, 1995), i.e.,

$$\mathbb{P}\left[\det G_k > \alpha^k\right] \leq \inf_{m \in \{1,2,\ldots\}} \frac{\mathbb{E}[\det G_k^m]}{\alpha^{km}}$$

$$\leq \inf_{m \in \{1,2,\ldots\}} \frac{\nu\left(\lambda_1^{(m)}, \lambda_2^{(m)}, \ldots\right)}{\alpha^{km}}.$$

However, analyzing $\lambda_i^{(m)}$ is difficult in general, and remains an open research question.

It is also possible to improve the second step, using concentration inequalities for configuration functions (Boucheron et al., 1999). Let $\psi_1, \ldots, \psi_k$ be a subsequence of $\phi_{1:n}$. We say $\psi_{1:k}$ is $\alpha$-*compatible*, if for

$j = 1, \ldots, k$,

$$\frac{\det G_{\{\psi_1,\ldots,\psi_j\}}}{\det G_{\{\psi_1,\ldots,\psi_{j-1}\}}} > \alpha.$$

Note that the dictionary constructed by OKS is $\alpha$-compatible, and the property of $\alpha$-compatibility is *hereditary*, i.e., $\psi_{1:k}$ being $\alpha$-compatible implies that all sub-sequences are also $\alpha$-compatible. To see this, let $\psi_{i_1}, \ldots, \psi_{i_s}$ be a sub-sequence of $\psi_{1:k}$, then

$$\begin{aligned}\frac{\det G_{\{\psi_{i_1},\ldots,\psi_{i_s}\}}}{\det G_{\{\psi_{i_1},\ldots,\psi_{i_{s-1}}\}}} &= \min_{\psi \in \text{span}\{\psi_{i_1},\ldots,\psi_{i_{s-1}}\}} \|\psi_{i_s} - \psi\|^2 \\ &\geq \min_{\psi \in \text{span}\{\psi_1,\ldots,\psi_{i_s-1}\}} \|\psi_{i_s} - \psi\|^2 \\ &= \frac{\det G_{\{\psi_1,\ldots,\psi_{i_s}\}}}{\det G_{\{\psi_1,\ldots,\psi_{i_s-1}\}}} > \alpha.\end{aligned}$$

As a result, let $Z_n$ denote the length of the longest sub-sequence in $\phi_{1:n}$ that is $\alpha$-compatible, then $|\mathcal{D}_n| < Z_n$. By Theorem 2 in Boucheron et al. (1999), $Z_n$ concentrates sharply around $\mathbb{E}[Z_n]$. Therefore, it is unlikely that $|\mathcal{D}_n|$ exceeds $\mathbb{E}[Z_n]$ by much. However, providing tight bounds for $\mathbb{E}[Z_n]$ is difficult and requires further study.

## References

F. R. Bach and M. I. Jordan. Kernel independent component analysis. *JMLR*, 3:1–48, 2002.

R. Bhatia and L. Elsner. The Hoffman-Wielandt inequality in infinite dimensions. *Proc. Indian Acad. Sci. (Math. Sci)*, 104(3):483–494, 1994.

G. Blanchard, O. Bousquet, and L. Zwald. Statistical properties of kernel principal component analysis. *Machine Learning*, 66:259–294, 2007.

S. Boucheron, G. Lugosi, and P. Massart. A sharp concentration inequality with applications. Technical Report 376, Department of Economics and Business, Universitat Pompeu Fabra, 1999.

M. L. Braun. *Spectral properties of the kernel matrix and their relation to kernel methods in machine learning*. PhD thesis, University of Bonn, 2005.

T. M. Cover and J. A. Thomas. Determinant inequalities via information theory. *SIAM J. Matrix Anal. Appl.*, 9(3):384–392, 1988.

P. Drineas and M. W. Mahoney. On the Nyström method for approximating a Gram matrix for improved kernel-based learning. *JMLR*, 6:2153–2175, 2005.

N. Duy and J. Peters. Incremental sparsification for real-time online model learning. In *AISTAT'10*, pages 557–564, 2010.

Y. Engel. *Algorithms and representations for reinforcement learning*. PhD thesis, Hebrew University, 2005.

Y. Engel, S. Mannor, and R. Meir. The kernel recursive least-squares algorithm. *IEEE Transactions on Signal Processing*, 52(8):2275–2285, 2004.

A. Gretton, K. Fukumizu, Z. Harchaoui, and B. K. Sriperumbudur. A fast, consistent kernel two-sample test. In *NIPS'09*, 2009.

L. Györfi, M. Kohler, A. Krzyzak, and H. Walk. *A distribution-free theory of nonparametric regression*. Springer, 2004.

Z. Harchaoui, F. R. Bach, and É. Moulines. Testing for homogeneity with kernel Fisher discriminant analysis. In *NIPS'08*, 2008.

W. Hoeffding. The strong law of large numbers for U-statistics. Technical Report 302, Department of statistics, University of North Carolina, 1961.

R. Jin, T.-B. Yang, M. Mahdavi, Y.-F. Li, and Z.-H. Zhou. Improved bound for the Nyström method and its application to kernel classification. Technical Report arXiv:1111.2262v3, 2012.

V. Koltchinskii and E. Giné. Random matrix approximation of spectra of integral operators. *Bernoulli*, 6(1):113–167, 2000.

J. Martin. The expected determinant of the random Gram matrix and its application to information retrieval systems, 2007. URL http://dydan.rutgers.edu/Seminars/Slides/martin2.pdf.

C. D. Meyer. *Matrix analysis and applied linear algebra*. SIAM: Society for Industrial and Applied Mathematics, 2001.

C. P. Niculescu. A new look at Newton's inequalities. *Journal of Inequalities in Pure and Applied Mathematics*, 1(2), 2000.

T. K. Philips and R. Nelson. The moment bound is tighter than Chernoff's bound for positive tail probabilities. *The American Statistician*, 49(2):175–178, 1995.

B. Schölkopf and A. J. Smola. *Learning with kernels: support vector machines, regularization, optimization, and beyond*. MIT Press, 2002.

R. J. Serfling. *Approximation theorems of mathematical statistics*. Wiley, 1980.

K. Slavakis, S. Theodoridis, and I. Yamada. Online kernel-based classification using adaptive projection algorithms. *Signal Processing, IEEE Transactions on*, 56(7):2781–2796, 2008.

I. Steinwart, D. Hush, and C. Scovel. An explicit description of the reproducing kernel Hilbert spaces of Gaussian RBF kernels. *IEEE Transactions on Information Theory*, 52:4635–4643, 2006.

C. Williams and M. Seeger. Using the Nyström method to speed up kernel machines. In *NIPS'00*, pages 682–688, 2000.

X. Xu. A sparse kernel-based least-squares temporal difference algorithm for reinforcement learning. In *Advances in Natural Computation*, volume 4221 of *Lecture Notes in Computer Science*, pages 47–56. Springer, 2006.